\documentclass{svproc}

\usepackage{graphicx} 
\usepackage{tabularray}
\usepackage{booktabs}
\usepackage{subcaption}
\usepackage{multirow}
\usepackage{xcolor}
\usepackage{amsmath}
\usepackage{amssymb}
\usepackage{dsfont}
\usepackage{bm}
\usepackage{array}
\usepackage{nicefrac, xfrac}
\usepackage[T1]{fontenc} 

\usepackage[noabbrev,capitalize,nameinlink]{cleveref}
\usepackage{url}

\usepackage[acronym]{glossaries}
\usepackage[per-mode=symbol]{siunitx}
\DeclareSIUnit{\degree}{°}
\DeclareSIUnit{\deg}{deg}
\DeclareSIUnit{\nothing}{\relax}
\DeclareSIUnit\pixel{px}
\definecolor{bgred}{HTML}{D11E1E}
\definecolor{bgblue}{HTML}{2E78E8}
\definecolor{bggreen}{HTML}{1ED130}
\definecolor{bgpurple}{HTML}{CA50DE}
\definecolor{bgyellow}{HTML}{E2D76F}

\newacronym[plural=SoCs, firstplural=systems-on-chip (SoCs)]{soc}{SoC}{System-on-Chip}
\newacronym{pid}{PID}{proportional-integral-derivative}
\newacronym[plural=PCBs, firstplural=printed circuit boards (PCBs)]{pcb}{PCB}{printed circuit board}
\newacronym{fc}{FC}{fabric controller}
\newacronym{fov}{FoV}{field of view}
\newacronym{cnn}{CNN}{convolutional neural network}
\newacronym[plural=LUTs, firstplural=lookup tables (LUTs)]{lut}{LUT}{lookup table}
\newacronym{uav}{UAV}{unmanned aerial vehicle}
\newacronym{tof}{ToF}{Time-of-Flight}
\newacronym{soa}{SoA}{State-of-the-Art}
\newacronym[plural=MCUs, firstplural=MicroController Units (MCUs)]{mcu}{MCU}{MicroController Unit}

\newcolumntype{L}[1]{>{\raggedright\let\newline\\\arraybackslash\hspace{0pt}}m{#1}}
\newcolumntype{C}[1]{>{\centering\let\newline\\\arraybackslash\hspace{0pt}}m{#1}}
\newcolumntype{R}[1]{>{\raggedleft\let\newline\\\arraybackslash\hspace{0pt}}m{#1}}

\makeatletter
\def\bstctlcite{\@ifnextchar[{\@bstctlcite}{\@bstctlcite[@auxout]}}
\def\@bstctlcite[#1]#2{\@bsphack
  \@for\@citeb:=#2\do{%
    \edef\@citeb{\expandafter\@firstofone\@citeb}%
    \if@filesw\immediate\write\csname #1\endcsname{\string\citation{\@citeb}}\fi}%
  \@esphack}
\makeatother

\begin{document}
\mainmatter              
\title{Combining Local and Global Perception for Autonomous Navigation on Nano-UAVs}

\titlerunning{  }  
%
\author{
Lorenzo Lamberti\inst{1} \and 
Georg Rutishauser\inst{2} \and 
Francesco Conti\inst{1} \and
Luca Benini\inst{1,2}
}
%
\authorrunning{Accepted for publication at the European Robotics Forum 2024.}
\titlerunning{Accepted for publication at the European Robotics Forum 2024.}

%
\tocauthor{Lorenzo Lamberti, Georg Rutishauser, Francesco Conti, Luca Benini}
%
\institute{DEI, University of Bologna, Italy,
\and
Integrated Systems Laboratory, ETH Z\"urich, Switzerland.
}
\maketitle              

\begin{abstract}
A critical challenge in deploying unmanned aerial vehicles (UAVs) for autonomous tasks is their ability to navigate in an unknown environment.
This paper introduces a novel vision-depth fusion approach for autonomous navigation on nano-UAVs.
We combine the visual-based PULP-Dronet~\cite{pulpdronetv2JETCAS} convolutional neural network for semantic information extraction, i.e., serving as the \textit{global perception}, with 8$\times$\SI{8}{\pixel} depth maps for close-proximity maneuvers, i.e., the \textit{local perception}.
When tested in-field, our integration strategy highlights the complementary strengths of both visual and depth sensory information. 
We achieve a 100\% success rate over 15 flights in a complex navigation scenario, encompassing straight pathways, static obstacle avoidance, and \SI{90}{\degree} turns. 

\end{abstract}

\bstctlcite{IEEEexample:BSTcontrol}


\section{Introduction}\label{sec:introduction}

With their sub-10 cm diameter and tens of grams in weight, autonomous nano-sized \glspl{uav} hold the potential to be employed in a wide range of applications, from inspection of hazardous environments~\cite{uav_chemicals} to warehouses~\cite{uav_industry_supplychain}.
Their small size enables access to confined spaces ~\cite{pulpdronetv2JETCAS,muller_2023_tof_matrix_avoidance}, and safe operation near humans~\cite{uav_industry_supplychain,palossi_tof_camera_fusion}, making them suitable for indoor environments.

Autonomous navigation is a pivotal requirement for any UAV exploring an environment while avoiding obstacles.
\gls{soa} navigation systems typically combine global and local planning techniques that are computationally heavy, relegating their execution to high-end devices such as GPUs~\cite{palossi_global_planner}. 
On the one hand, global planning is in charge of setting high-level destination goals, for example, ensuring the UAV is passing through pre-defined checkpoints.
On the other hand, local planning handles close proximity maneuvers, like obstacle avoidance, ensuring progress toward the goals set by the global planner.

The miniaturized form factor of nano-drones imposes significant constraints on their onboard computing devices, ultimately relying on low-power \glspl{mcu} with a power envelope of $\sim$\SI{100}{\milli\watt}~\cite{pulpdronetv2JETCAS}.
Consequently, \gls{soa} navigation algorithms on autonomous nano-\gls{uav} rely on computationally affordable solutions rather than complex planning strategies.
The exploration policies deployed by \cite{kimberly_bug_inspired,lamberti_exploration_detection} have to rely on basic state machines and single-beam VL53L1x \gls{tof} sensors for collision avoidance.
However, the limited \gls{fov} (\SI{15}{\degree}) of this \gls{tof} sensor challenges the detection of narrow obstacles~\cite{mahyar_coll_avoid}.
M\"uller \textit{et al.}~\cite{muller_2023_tof_matrix_avoidance} developed a depth-based navigation system for nano-UAVs, utilizing a novel multizone \gls{tof} sensor and implementing a lightweight decision tree based on 8$\times$\SI{8}{\pixel} depth maps.
While \gls{tof} sensors offer geometrically precise measurements of three-dimensional objects in the environment, they lack semantic information.
Visual-based navigation approaches, such as the \gls{soa} PULP-Dronet~\cite{pulpdronetv2JETCAS} \gls{cnn}, are better equipped to process semantically rich non-volumetric visual cues, such as lanes on a floor, that are invisible to \gls{tof} sensors.
On the other hand, these methods do not extract any precise geometric information about objects in the \gls{fov}, posing challenges for obstacle avoidance. 
While some research has explored the synergy between vision and depth sensors on nano-drones~\cite{palossi_tof_camera_fusion,mahyar_coll_avoid}, their focus has not been on autonomous navigation applications.

In this work, we take one step towards bringing the global and local planning paradigm aboard nano-\glspl{uav}.
We present a pipeline for autonomous navigation on nano-drones that integrates both \textit{global} and \textit{local} perception. 
The \textit{global perception} pipeline utilizes the PULP-Dronet \gls{cnn} for semantic information extraction, while the \textit{local perception} outputs close-proximity ($\sim$\SI{4}{\meter}) obstacle-free navigation area.
We fuse the visual-depth information with a lightweight lookup table approach, which maps all possible combinations of outputs from the local and global perception pipelines to drone control commands.

Our fused perception pipeline achieved a 100\% success rate on a set of fifteen flights, successfully navigating through straight pathways, executing  \SI{90}{\degree} turns, and avoiding static obstacles. 
In the same scenario, using only the global perception pipeline led to failures in static obstacle avoidance.
Conversely, when relying solely on the local perception pipeline, the system failed to handle 90° turns due to the absence of semantic features in the input data.
Our results highlight the benefit of combining depth and vision sensory inputs to enhance nano-UAV navigation while also revealing the limitations of depending solely on one sensor type in complex navigation scenarios.
\section{System design} \label{sec:methodology}

\begin{figure}[tb]
    \centering
    \includegraphics[width=1\linewidth]{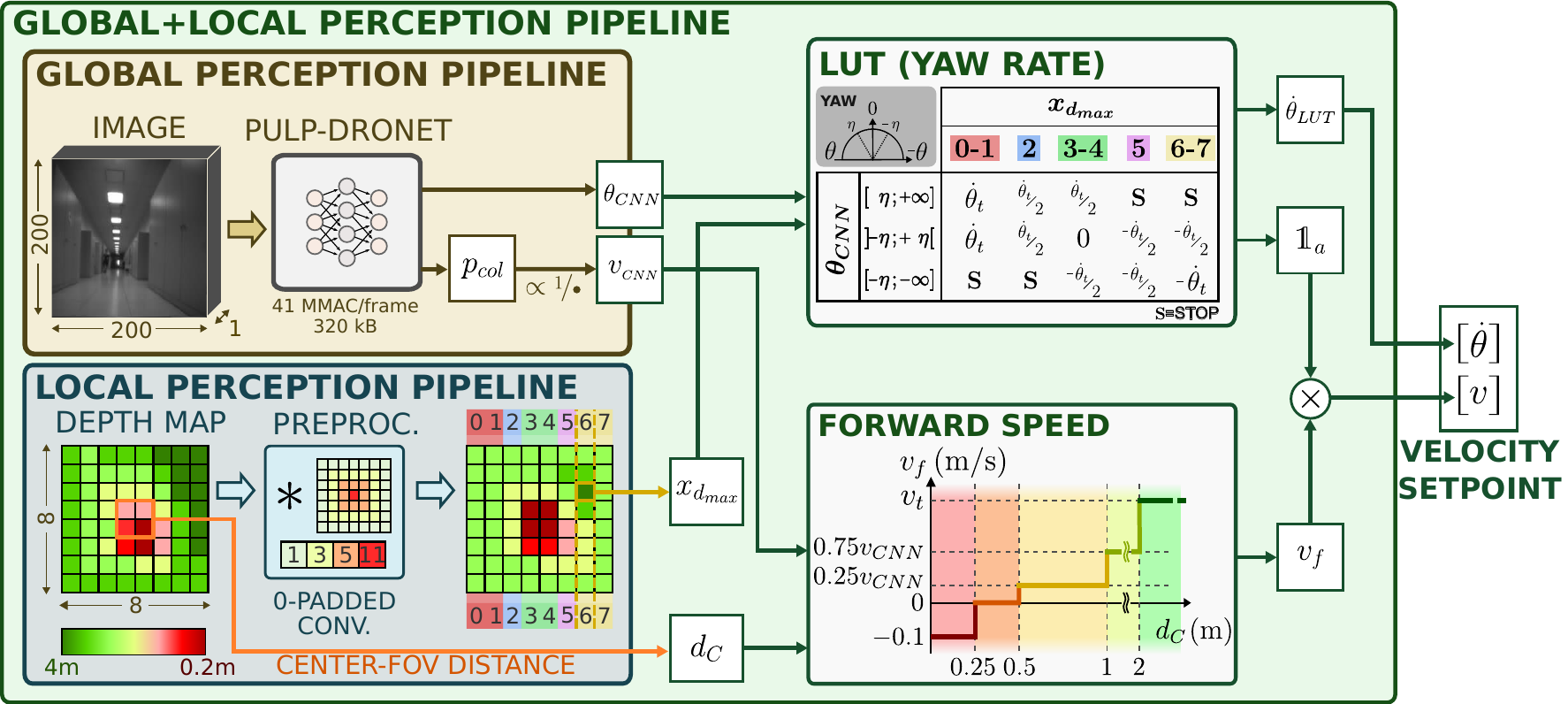}
    \caption{Our global $+$ local perception pipeline.}
    \label{fig:perception_pipeline}
    \vspace{-7pt}
\end{figure}

\textbf{Robotic platform.} 
We use the Bitcraze Crazyflie 2.1, a 27-gram, 10-\SI{}{\centi\meter} diameter nano-drone.
The onboard STM32 \gls{mcu} processor manages sensor interfacing and low-level flight control tasks. 
We extend the platform with three \glspl{pcb}.
The \emph{AI-deck} serves as the visual navigation engine, featuring a GWT GAP8 9-core RISC-V-based \gls{soc} and a grayscale QVGA Himax HM01B0 camera with a \SI{115}{\degree} \gls{fov}.
In the GAP8 architecture, a single core handles system operations,
including UART communication with the STM32 \gls{mcu},
while the general-purpose 8-core cluster is used for \gls{cnn} processing.
We use the open-source \textit{Multizone Ranger Deck}~\cite{muller_2023_tof_matrix_avoidance}, featuring a front-looking VL53LC5CX1 matrix \gls{tof} sensor. 
This sensor, connected to the STM32 over the I2C bus, acquires 8$\times$\SI{8}{\pixel} depth maps with a \SI{65}{\degree} \gls{fov} at a frequency of \SI{15}{\hertz} and with a range of \SI{0.2}{\nothing}--\SI{4}{\meter}.
Last, the \textit{Flow deck v2}, a downward-facing \gls{pcb} that provides the STM32 \gls{mcu} with height measurements from the ground and horizontal motion estimation based on optical flow.

\textbf{Global $+$ local perception pipelines.}
Our autonomous navigation pipeline for nano-\glspl{uav}, depicted in \cref{fig:perception_pipeline}, comprises several components, all executed onboard.
The \emph{global perception pipeline} is constituted by the PULP-Dronet~\cite{pulpdronetv2JETCAS}, a \gls{cnn} for visual-based navigation on nano-\glspl{uav}.
We use the pre-trained weights of~\cite{pulpdronetv2JETCAS}.
The \gls{cnn} takes $200\times200$\SI{}{\pixel} images as input and produces two outputs: a steering angle $\theta_{CNN}$ and a collision probability $p_{col}$.
This \gls{cnn} runs on the 8-core cluster of the GAP8 \gls{soc} at \SI{19}{FPS}.
As in~\cite{pulpdronetv2JETCAS}, we post-process the PULP-Dronet outputs, computing the forward speed $v_{CNN}$ inversely proportional to $p_{col}$, while the steering angle $\theta_{CNN}$ is rescaled into a yaw-rate.

The \emph{local perception pipeline} executes entirely on the STM32 \gls{mcu}.
We acquire the 8$\times$\SI{8}{\pixel} depth map from the front-looking matrix \gls{tof} sensor.
Then, we convolve the depth map with a 2-D Gaussian kernel, requiring only \SI{1600}{MAC} operations, to identify the most free area within the input.
The input is zero-padded to maintain consistent input-output dimensions and to favor navigation toward the center of the \gls{tof} \gls{fov} when no obstacles are present.
Finally, we determine the x-coordinate ($x_{d_{max}}$) of the pixel with the longest distance in the 8x8 depth map.
This coordinate ranges from 0 to 7, denoting a left turn in the 0--2 range, a right turn in the 5--7 range, and no turn otherwise.

To build our final navigation pipeline, we fuse the outputs coming from the global $+$ local perception pipelines.
PULP-Dronet's output $\theta_{CNN}\in [-1,1]$ is discretized with a threshold $\eta=0.1$.
We use a \gls{lut} to generate output pairs $(\mathds{1}_a, \dot{\theta}_{LUT} )$ for every possible combination of $(\theta_{CNN}, x_{d_{max}})$ inputs.
Here, $\dot{\theta}_{LUT}$ denotes the output yaw rate, while $\mathds{1}_a$ is a binary agreement indicator that sets the drone's forward speed: it assumes a value of 0 when $\theta_{CNN}$ and $x_{d_{max}}$ indicate opposite steering directions, as represented by $\mathbf{S}$ in the \gls{lut}, or it is 1 otherwise.
The output yaw-rate can be either zero, indicate left and right turns at the maximum target yaw rate ($\dot{\theta}_{t}$ for left, $-\dot{\theta}_{t}$ for right), or indicate left and right turns at half of the maximum target yaw rate ($\nicefrac{\dot{\theta}_{t}}{2}$ and $-\nicefrac{\dot{\theta}_{t}}{2}$, respectively).
The forward speed of the drone $v_f$, which is gated by $\mathds{1}_a$, is adjusted in step increments as depicted in \cref{fig:perception_pipeline}, and depends on the average distance ($d_C$) measured by the four central pixels of the \gls{tof} sensor.
To test the global perception pipeline in isolation, we use the \gls{cnn} outputs as in ~\cite{pulpdronetv2JETCAS}, and for isolated testing of the local perception pipeline, we force the \gls{cnn} outputs to $p_{col}=0$ and $\theta_{CNN}=0$.

\section{Results} \label{sec:results}

\begin{figure}[tb]
    \centering
    \includegraphics[width=1\linewidth]{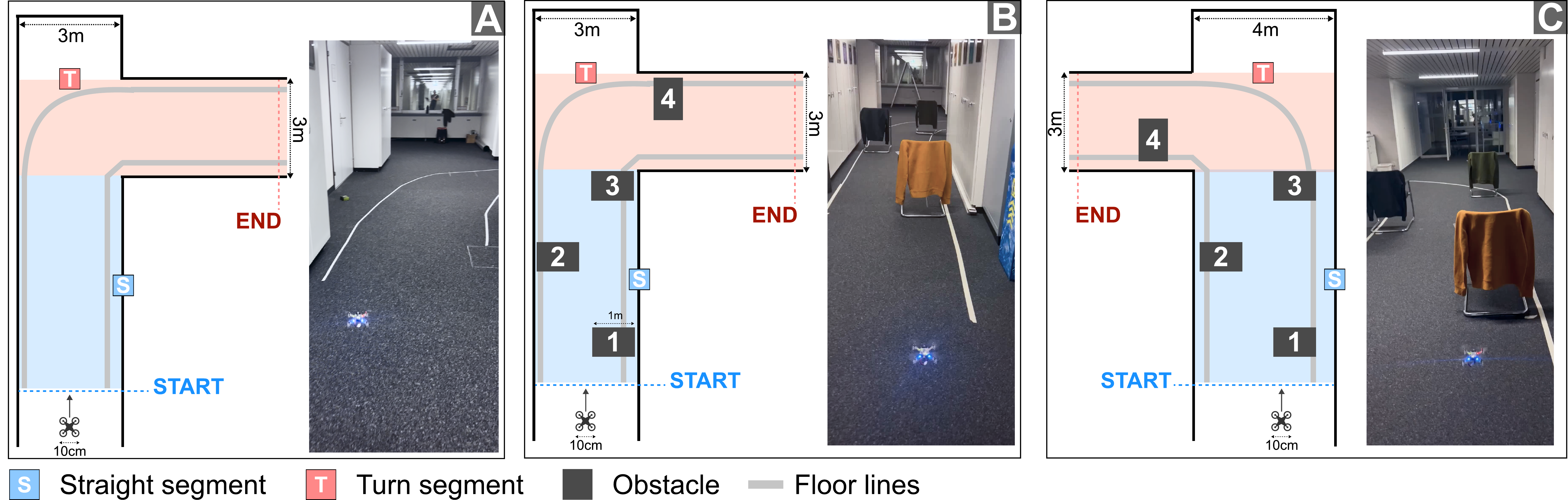}
    \caption{The testing setups for scenario 1 (A), scenario 2 (B), and scenario 3 (C).}
    \label{fig:testing_setup}
    \vspace{-7pt}
\end{figure}

We assess the navigation capabilities in an office corridor with a width of 3--\SI{4}{\meter} (\cref{fig:testing_setup}).
We set up three testing scenarios, each comprising a \textit{straight} pathway followed by a 90-degree \textit{turn}, highlighted in \cref{fig:testing_setup} with blue and red colors, respectively. 
Scenarios 1 and 2 entail a right turn (A and B), with the former obstacle-free and the latter containing four \SI{1}{\meter}-wide obstacles. 
Scenario 3 (C) mirrors scenario 2, featuring a left turn and four \SI{1}{\meter}-wide obstacles.
We draw white lines on the floor, resembling the car lanes of the PULP-Dronet training dataset~\cite{pulpdronetv2JETCAS}.
We assess the performance of the three perception pipelines discussed in \cref{sec:methodology} by conducting five flight tests per scenario, totaling 45 flights.
We set the drone's target height to \SI{0.5}{\meter}, the target forward speed ($v_t$) to \SI{1.5}{\meter/\second}, and $\dot{\theta}_{t}$ to \SI{60}{\deg/\s}.
A test is considered successful when the drone successfully flies through the corridor from \texttt{START} to \texttt{END} without collisions.

\cref{tab:results_static_and_dynamic_obstacles} reports the success rate of the drone for each section. 
Examples of the flight tests are available in the supplementary video\footnote{\url{https://youtu.be/J703fo_zIKQ}}.
In obstacle-free scenario 1, the global perception pipeline succeeds \SI{100}{\percent} of the time, successfully recognizing the visual clues on the floor.
However, this perception pipeline never succeeds in scenarios 2 and 3, colliding with the obstacles in every trial.
These results confirm that PULP-Dronet v2 struggles to tackle static obstacle avoidance, as mentioned in~\cite{pulpdronetv2JETCAS}.
Conversely, the local perception pipeline reliably avoids static obstacles with a 100\% success rate in straight segments.
Nonetheless, it consistently fails to execute turns in all scenarios, ultimately getting stuck at the end of the corridor, as the \gls{tof} can not recognize semantic elements of the environment, e.g., the floor's line markings.

Finally, we test our fused global $+$ local perception pipeline, which scores a \SI{100}{\percent} success rate across both the straight and turn segments of the corridor, both when navigating in an obstacle-free corridor (scenario 1) and when tackling obstacle avoidance (scenario 2 and 3).
In conclusion, our fused global $+$ local perception pipeline captures the benefits of both the depth-based and vision-based sensory inputs: the \gls{tof} ranging measurements allow for reliable short-range static obstacle avoidance, while the PULP-Dronet understands semantic cues from its visual input, allowing the drone to follow the shape of the corridor.

\begin{table}[tb]
\centering
\footnotesize
\caption{The success rate of the three perception pipelines on three scenarios.}
\label{tab:results_static_and_dynamic_obstacles}
\begin{tabular}{L{20mm}R{16.5mm}R{12.5mm}R{16.5mm}R{12.5mm}R{16.5mm}R{12.5mm}}
\toprule
\textbf{Perception} &   \multicolumn{2}{c}{\hphantom{aaa}\textbf{Global}}   & \multicolumn{2}{c}{\hphantom{aaaa}\textbf{Local}}   & \multicolumn{2}{c}{\hphantom{aaa}\textbf{Global $+$ Local}}  \\
Section &    Straight &  Turn\hspace{0.35mm}\hphantom{\,}&  Straight &  Turn  &  Straight &  Turn\hspace{0.35mm}\hphantom{\,} \\\midrule
Scenario 1 &    \SI{100}{\percent}\hphantom{i\,} &  \SI{100}{\percent} &  \SI{ 100}{\percent}\hphantom{i\,}&  \SI{0}{\percent}\hphantom{\,} &  \SI{100}{\percent}\hphantom{i\,} & \SI{100}{\percent}\\
Scenario 2 &    \SI{0}{\percent}\hphantom{i\,} &  N/A &   \SI{100}{\percent}\hphantom{i\,}&  \SI{0}{\percent}\hphantom{\,}&   \SI{100}{\percent}\hphantom{i\,} & \SI{100}{\percent}\\
Scenario 3 &   \SI{0}{\percent}\hphantom{i\,} &  N/A &   \SI{100}{\percent}\hphantom{i\,} &  \SI{0}{\percent}\hphantom{\,}&   \SI{100}{\percent}\hphantom{i\,} & \SI{100}{\percent} \\\bottomrule
\end{tabular}
\vspace{-7pt}

\end{table}

\section{Conclusion} \label{sec:conclusion}

We presented a pipeline for autonomous navigation on nano-\glspl{uav} that combines global and local perception. 
The global perception exploits the visual-based PULP-Dronet \gls{cnn} to extract visual cues from the input images, while the local perception relies on an 8$\times$\SI{8}{\pixel} Time-of-Flight (\gls{tof}) sensor capturing reliable close-proximity occupancy maps.
Our fused perception pipeline achieved a 100\% success rate in fifteen flights, navigating through straight pathways, avoiding static obstacles, and executing a \SI{90}{\degree} turn.


\section*{\normalsize{Acknowledgments}}
\vspace{-5pt}
\small We thank D. Palossi and D. Christodoulou for their contribution to this work. 


\bibliographystyle{IEEEtran}

\bibliography{IEEEabrv,\jobname,author}

\end{document}